\documentclass[lettersize,journal]{IEEEtran}

\usepackage{graphicx}
\usepackage{amsmath}
\usepackage{amssymb}
\usepackage{cite}
\usepackage[letterpaper,top=54pt,bottom=54pt,left=54pt,right=54pt]{geometry}
\usepackage{epstopdf}
\usepackage[lofdepth,lotdepth]{subfig}
\usepackage{booktabs}
\usepackage{color,soul}

\IEEEoverridecommandlockouts
\title{\bf BRL/Pisa/IIT SoftHand:
	A Low-cost, 3D-Printed, Underactuated, Tendon-Driven Hand with Soft and Adaptive Synergies 
	\vspace{0em}
}

\author{Haoran Li, Christopher J. Ford, Matteo Bianchi, Manuel G. Catalano, Efi Psomopoulou, Nathan F. Lepora
	\thanks{HL was supported by the the China Scholarship Council. NL and EP were supported by a Leadership Award from the Leverhulme Trust on ‘A biomimetic forebrain for robot touch’ (RL-2016-39).}
	\thanks{HL, CF, EP and NL are with the Department of Engineering Mathematics and Bristol Robotics Laboratory, University of Bristol, Bristol, U.K. (e-mail: haoran.li@bristol.ac.uk, chris.ford@bristol.ac.uk,  efi.psomopoulou@bristol.ac.uk, n.lepora@bristol.ac.uk). \newline \indent MB is with the Department of Information Engineering and the Research Center "E.Piaggio", University of Pisa, Italy (e-mail: matteo.bianchi@centropiaggio.unipi.it). \newline \indent MGC is with the Soft Robotics for Human Cooperation and Rehabilitation, Istituto Italiano di Tecnologia (IIT), Italy (e-mail: manuel.catalano@iit.it.)}
	\thanks{Project repository: https://github.com/SoutheastWind/BPI-SoftHand}
	}

\begin{document}
	
	\maketitle
	\thispagestyle{empty}
	\pagestyle{empty}
	
	%%%%%%%%%%%%%%%%%%%%%%%%%%%%%%%%%%%%%%%%%%%%%%%%%%%%%%%%%%%%%%%%%%%%%%%%%%%%%%%%
	\begin{abstract}
	This paper introduces the BRL/Pisa/IIT (BPI) SoftHand: a single actuator-driven, low-cost, 3D-printed, tendon-driven, underactuated robot hand that can be used to perform a range of grasping tasks. Based on the adaptive synergies of the Pisa/IIT SoftHand, we design a new joint system and tendon routing to facilitate the inclusion of both soft and adaptive synergies, which helps us balance durability, affordability and \textcolor{black}{grasping performance} of the hand. The focus of this work is on the design, simulation, synergies and grasping tests of this SoftHand. The novel phalanges are designed and printed based on linkages, gear pairs and geometric restraint mechanisms, and can be applied to most tendon-driven robotic hands. We show that the robot hand can successfully grasp and lift various target objects and adapt to hold complex geometric shapes, reflecting the successful adoption of the soft and adaptive synergies. We intend to open-source the design of the hand so that it can be built cheaply on a home 3D-printer. 
	\\
	For more detail:
	https://sites.google.com/view/bpi-softhand-tactile-group-bri/brlpisaiit-softhand-design
	%新颖的指骨是基于连杆，齿轮副以及几何约束机构设计并打印，可被应用大多数tendon-driven机器手。
	
	\end{abstract}

	%%%%%%%%%%%%%%%%%%%%%%%%%%%%%%%%%%%%%%%%%%%%%%%%%%%%%%%%%%%%%%%%%%%%%%%%%%%%%%%%
	\section{INTRODUCTION}
	%\vspace{-1pt}
	The human hand is highly capable at grasping due to its mechanical structure and dexterity. The embodiment of its grasping ability is based on the synergistic operation of the ligaments, tendons and muscles underlying the operation of human hands. The concept of a soft synergy was proposed by Bicchi et al~\cite{bicchi2011modelling}, and a transmission scheme based on this concept can address the problem of matching the actual internal force required by the controlled object with the force generated by the system. However, its mechanical structure is difficult to realize on robotic hands. Catalano et al.\cite{catalano2014adaptive} proposed the concept of an adaptive synergy and designed the Pisa/IIT SoftHand based on this principle applied to soft anthropomorphic robot hands. Its experiments show that Pisa/IIT SoftHand performs well in grasping experiments and manipulation experiments utilizing adaptive synergies~\cite{catalano2014adaptive}.

    Over the past few years, there has been a steady increase in interest in designing and building robotic hands. However, there is still just a handful of inexpensive, 3D-printed bionic underactuated robotic hands that can apply the synergy concept\cite{piazza2019century}.	In this study, we focus on how to effectively and concisely apply the concepts of both adaptive synergies and soft synergies to a low-cost, 3D-printed robotic hand. 
    
    Our new robotic hand draws from the tendon-layout design technology of the Pisa/IIT SoftHand on its joints, which we supplement with new joint structures and passive reduction ligaments. The adaptive synergy is instantiated between the robotic fingers due to their coupling via the tendons. The tendon-layout design technology is inspired from the open-source model of Pisa/IIT SoftHand~\cite{della2017quest}. The original Pisa/IIT had just an adaptive synergy, which works effectively thanks to a fine balance between friction and elastic ligaments. However, here we propose a lower-cost, 3D-printed SoftHand, and so also take 
    advantage of a novel soft synergy design scheme to reduce the loss of force in the transmission, which is to add a spring between each tendon and the driver. 

\begin{figure}[t]
		\centering
		\begin{tabular}{@{}c@{}}
			\includegraphics[width=0.9\columnwidth,trim={0 0 0 0},clip]{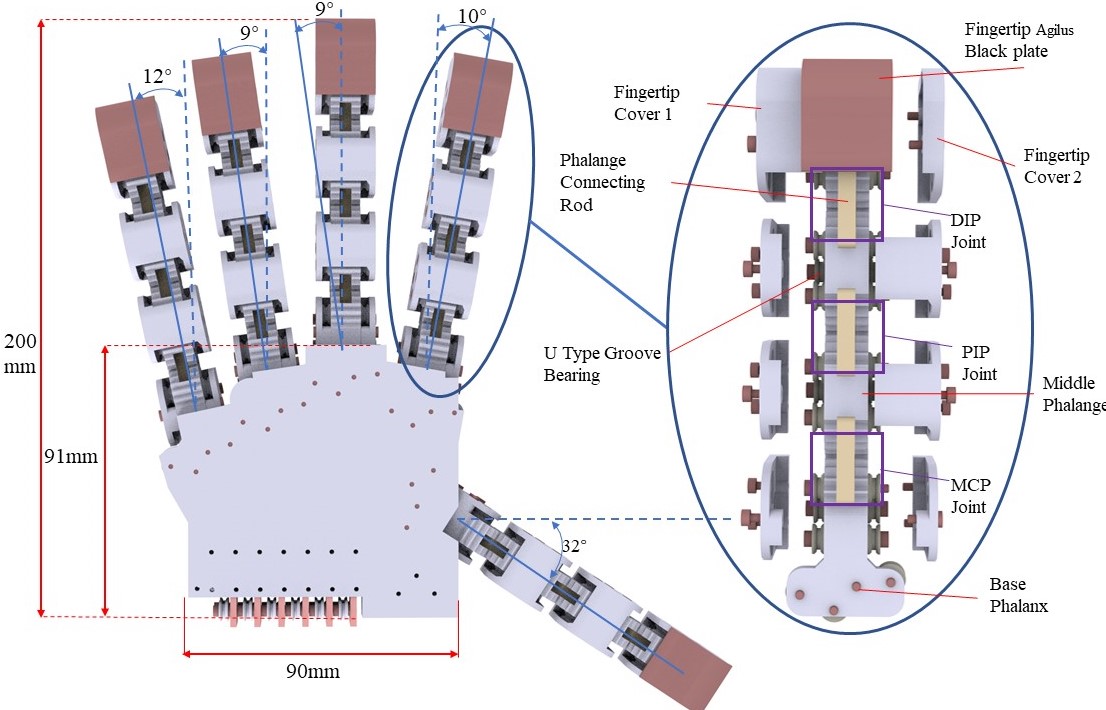}
		\end{tabular}
% 		\vspace{-5pt}
		\caption{The design overview of the BRL/Pisa/IIT SoftHand, indicating modular components and general dimensions. The right exploded figure shows the main components of a single modular finger. The purple boxes represent the three finger joints (see Fig.~\ref{fig2} for more details).}
% 		\vspace{-5pt}
		\label{fig1}	
	\end{figure}

    Our main contributions in this work are to design and build a low-cost, 3D-printed, underactuated, tendon-driven softhand with both soft and adaptive synergies:\\
	\noindent 1) We designed and built a novel low-cost, single-drive SoftHand that can be fabricated on a home 3D-printer with a new joint system and tendon routing, which we term the BRL/Pisa/IIT SoftHand (Fig.~\ref{fig1}).\\
	\noindent 2) We applied soft synergy and adaptive synergy in the BRL/Pisa/IIT softhand and verified its grasping ability through two experiments that assess finger closure onto an object and effectiveness of holding that object.

   This paper is organized as follows. Section II briefly describes some representative underactuated robot hands. Section III introduces the detailed design of our SoftHand and its key components. Section IV focuses on the robotic hand's workspace through kinematic simulation which are used to evaluate grasping, dexterity and improve the design. Section V reports on grasping experiments that verify the SoftHand's synergy ability and grasping ability. Section VI concludes the paper and discusses future work.

	%%%%%%%%%%%%%%%%%%%%%%%%%%%%%%%%%%%%%%%%%%%%%%%%%%%%%%%%%%%%%%%%%%%%%%%%%%%%%%%%
	\section{BACKGROUND AND RELATED WORK} \label{bgd}

	Using the latest fabrication techniques and technology to endow robotic hands with human-like sensory, dexterous and grasping capabilities remains one of the greatest challenges in engineering \cite{piazza2019century}\cite{9499032}. \textcolor{black}{Robot hands have imitated human perception by integrating various sensors\cite{james2021tactile}\cite{lepora2021towards} that can provide the interaction forces and surface properties of the interface between the objects and robot hand\cite{kappassov2015tactile}\cite{9499032}, and imitating the movement of joints with actuators\cite{butterfass2001dlr}\cite{fan2018research}\cite{alspach2018design}} Because fully-actuated robotic hands\cite{martin2014design}\cite{palli2014dexmart} are not widely available due to the difficulty of controlling them and their high fabrication cost, more researchers are focusing on designing robotic hands with fewer actuators but relatively high anthropomorphic functions\cite{piazza2019century}. Lightweight, compact and tendon-driven underactuated schemes have becomes a principal option for functional robotic hands\cite{jiang2014modular}\cite{catalano2014adaptive}\cite{piazza2019century}\cite{xiong2016design}. One can also integrated tendons and underactuated robotic fingers as modules independent of other components~\cite{liow2019olympic,jiang2014modular}, which eases user maintenance but also makes the palm structure more complex. Soft materials can be used in the underactuated robotic hand \cite{mizushima2018multi} with springs between the tendons and the actuator, which greatly improved the adaptability and grasping ability of the fingers. 
	
	The Pisa/IIT SoftHand\cite{catalano2014adaptive} uses a single motor to drive all joints and has good adaptability to a wide range of objects. \textcolor{black}{Chen et al.\cite{chen2014mechanical} extracted the motion synergy, in the form of the angular velocity relationship of different joints, from a large number of grasping tasks. The X-hand \cite{xiong2016design} uses four DC motors to control the joint-coupled fingers and motion transmission mechanism. It performs well in experiments on grasping objects of different shapes by changing the posture of the thumb and the coordination of the remaining four fingers. Sun et al.\cite{sun2021design} has improvements the X-hand\cite{xiong2016design}, designed a mini-hand whose thumb is independent of the four-finger movement and the four fingers cooperate through a single-actuated differential palm mechanism.} %  It can achieve 97\% human grasping characteristics among four fingers by using two DC motors.
	
	However, it is a challenging and time-consuming task to manufacture low-cost, easy-to-maintain and user-customizable tendon-driven robotic hands. In the present paper, we use the combined drive scheme of both soft and adaptive synergies, aiming to find a compromise solution to balance the complexity of the robot hand, the difficulty of design and manufacture, and the grasping capability. \\
	%%%%%%%%%%%%%%%%%%%%%%%%%%%%%%%%%%%%%%%%%%%%%%%%%%%%%%%%%%%%%%%%%%%%%%%%%%%%%%%%	
    
    \section{Development of BRL/Pisa/IIT SoftHand}\label{Development}
	In this section, we describe the design of a novel, underactuated robot hand according to the design methods of adaptive synergies and soft synergies. The functional requirement is that the new robotic hand can grab common objects in daily life, such as screwdrivers and water bottles. The manufacturing design requirements are that the structure should be simple, reliable and easy to maintain. It also should be friendly and easy to control in terms of human-robot hand interaction. The overall parameters and design details of the relevant components are detailed below.
	
	  \begin{figure}[t]
		\centering
		\begin{tabular}{@{}c@{}}
			\includegraphics[width=0.95\columnwidth,trim={0 0 0 0},clip]{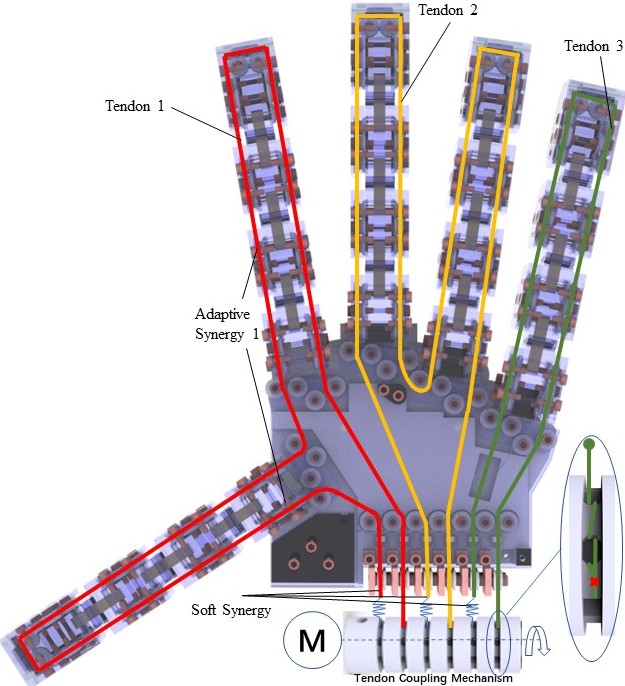}
		\end{tabular}
		\caption{Tendon Layout: The left tendon (in red) routes motion to the joints of thumb and index finger; the middle  tendon (yellow) routes motion to the joints of middle and third fingers; the right tendon (green) routes motion to the joints of just the little (pinky) finger.}
		\vspace{-1em}
		\label{fig4}	
	\end{figure}

	\subsection{BRL/Pisa/IIT SoftHand Overall Design}
	
	The hand has 15 degrees of freedom and actuated by one Maxon DC motor (347486). It uses 3 different tendons that pass through distinct groups of fingers and the thumb (Fig.~\ref{fig4}). These three tendons pass through the fingers and palm through U-type groove guiding bearings, and are coupled by the tendon coupling mechanism at the end of the palm. The overall size, finger layout, their angles and modular finger components are shown in Fig.~\ref{fig1}
	
	In its rest position, the overall length of the SoftHand is 200mm, measured from the palm base to the fingertip of the middle finger. The span from the tip of the thumb to the tip of the little finger is 215mm. The width and length of palm are 90mm and 91mm respectively. The angles between fingers and the vertical line to the palm base are $10^\circ$,  $0^\circ$, $9^\circ$, $12^\circ$ from right to left starting from index finger. The angle between the thumb and the horizontal line to the palm is $32^\circ$. These values were carefully tuned because the angles between the under-actuated fingers greatly influences on gripping space, especially where the fingertips are located. The above-designed finger angles are based on simulation experiments of robotic gripping kinematics and human hand bionics. We will verify its grasping ability later by kinematics simulation and grasping experiments. 
	
    Except for the U-type groove bearing (Figs~\ref{fig2},\ref{fig3}), the entire structure of the BPI SoftHand can be made with a home-use 3D-printer. To increase the friction between the finger contact surface and the grasped object, each fingertip is equipped with a 1.5mm thick Agilus Black material surface (see Fig.~\ref{fig1}, indicated in brown), although other materials to provide traction upon grasping would also be suitable (e.g. high-friction tape). The user can choose a suitable drive motor according to the torque of the motor. Excluding the motor, the overall materials cost of building the hand is around £65.
    
	\subsection{Phalanx and Fingertip Design}
	\subsubsection{Modular Finger Design}
	All fingers and other parts of the robot hand are independent of each other, and fingers are also of the same design. This means that any robot finger can be easily replaced without affecting other hand structures, which greatly improves the ease of maintenance. The length of the standard finger is 115mm that measured from its base phalange to its fingertip, and its width is 20mm (See the right part of Fig.\ref{fig1}).  The fingertip, middle phalanges and all phalanges cover were 3D printed in PLA. In order to obtain better mechanical properties, the base phalanx used in this paper were 3D-printed using ABS.
	
\begin{figure}[t]
		\centering
		\begin{tabular}{@{}c@{}}
			\includegraphics[width=1\columnwidth,trim={0 0 0 0},clip]{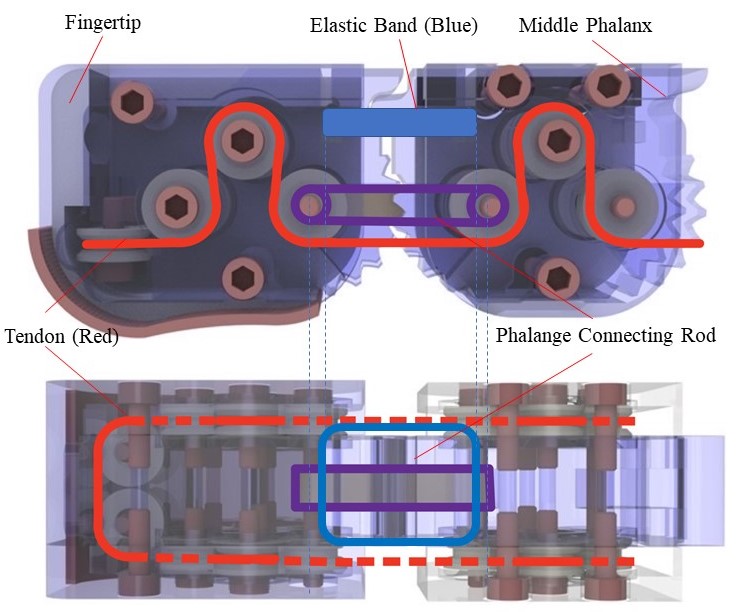}
		\end{tabular}
		\vspace{-5pt}
		\caption{Side and top views of the modular fingertip and middle phalanx, showing the tendon layout (red line), elastic band (blue box) and phalanx connecting rod (purple box).}
		\vspace{-5pt}
		\label{fig2}	
        \vspace{1em}
		\begin{tabular}{@{}c@{}}
			\includegraphics[width=1\columnwidth,trim={0 0 0 0},clip]{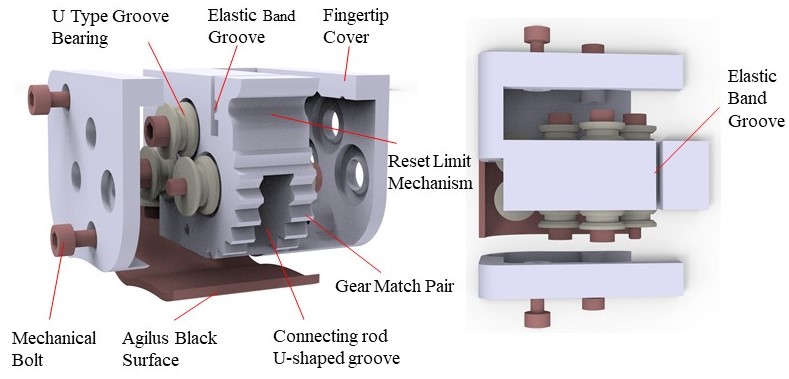}
		\end{tabular}
		\caption{Exploded view of a single fingertip.}
		\vspace{-1em}
		\label{fig3}	
	\end{figure}

    The modular fingers each consist of four modular phalanges, including a fingertip, two mid-phalanges and a base phalanx. Each phalanx has two phalanx covers and several U-type grooved bearings used to guide the tendon and decrease the friction between the tendon and phalanx. Several mechanical bolts with different lengths (from 6mm to 18mm) are used to connect the phalange cover, U-type bearing and phalange connecting rod (purple region in Fig.\ref{fig2}).

	Each finger has 3 degrees of freedoms including: metacarpophalangeal (MCP), proximal interphalangeal (PIP), and distal interphalangeal (DIP) joints that allow flexion and extension of the fingers (see purple boxes to right-side of Fig.\ref{fig1}). To reduce the complexity of controlling the hand and to improve the degree of modularity, the carpometacarpal (CMC) joint of the thumb is replaced by metacarpophalangeal (MCP) joint. Hence, the modified thumb consists of MCP, PIP and DIP joints. \textcolor{black}{ The CMC, DIP, PIP phalanges are connected by phalanx connecting rods (purple box on Fig.~\ref{fig2}) to limit the longitudinal and lateral degrees of freedom of the phalanx.} The relative rotation between the phalanxes is achieved through gear pairs (Fig.\ref{fig3}). Due to the modular design, the transmission ratio between gear pairs can be adjusted by replacing the related phalanges to obtain different rotation angles and grasping postures. Furthermore, the modular phalanx and finger design provides opportunities to realize various geometric shapes, materials, and tendon routing paths to achieve different functions for the hand.
	 
	\subsubsection{Middle Phalanx Design} \label{pinch}
	There are three U-type bearings on each side of the middle phalanx. The bearing layout is inspired by Pisa/IIT SoftHand \cite{catalano2014adaptive}. A tendon passes through these three bearings in turn on one side, then returns on three opposing bearings on the other side (tendon layout shown in Fig.\ref{fig2}). The relative positional relationship between these three bearings determines the friction between the tendons and these bearings and the torque to rotate the phalanx. After assessing the friction through adjusting the relative positional relationship of the three bearings, the proposed layout offers appropriate rotational torque, friction and size. In addition, 2 grooves on the upper part of the middle phalanx hold an elastic band for the resetting movement (Fig.~\ref{fig2}). The first groove is used to hold the elastic band from the first phalanx and the latter groove holds the elastic band on the second phalanx \textcolor{black}{(rubber of stiffness $\sim$0.49N/mm)}. Therefore, under the action of the elastic force of the anterior and posterior elastic bands, an appropriate relative rotation occurs between the phalanges to complete the reset motion.

	\subsubsection{Modular Fingertip Design} \label{pinch}
	Compared with the middle phalanx, the fingertip has two extra U-type grooved bearings that guide the tendon to return, which allows the tendon to enter from one side of the finger and leave from the other side (see Fig.~\ref{fig3}). A rubber-like soft material coating (here made from Agilus black) is included to increase the friction between the fingertip and the grasped object.
	
	\subsubsection{Base Phalanx Design} \label{pinch}
	The base phalanx is not only used to fix the modular fingers to the palm, but also acts to guide how the tendon enters and leaves the finger (see right of Fig.~\ref{fig1}). The base plate also has a guide bearing to avoid direct contact between tendon and phalanx, which reduces friction. The rest of its structural design is similar to the middle phalanx.
	
	\subsection{Palm Design and Tendon Layout}\label{Layout}

	Due to being tendon driven, in this design the motor and control board can be placed outside the robot hand. Hence, the palm and fingers are purely mechanical with no electronics. In principle, the design could be customized to house the motor within the palm, as in the Pisa/IIT SoftHand. 	
	
	As shown in Figure \ref{fig4}, three independent tendons pass through the five fingers, grouped into the thumb and index fingers, middle and third fingers, and little (pinky) finger. All tendons are coupled with a \textcolor{black}{coupling} mechanism at the base of the palm. \textcolor{black}{The tendon ends pass through distinct channels of the tendon coupling mechanism and are fixed in place by knotting. All tendons are driven by a single actuator. }
	
	\textcolor{black}{We chose this tendon configuration for three reasons. Firstly, the inclusion of both soft and adaptive synergies needs at least 2 independent tendons. Secondly, the purpose of using a third tendon to drive one finger alone is for finger synchronization, because using one tendon to drive more than two independent fingers can result in lag in the middle finger movement. Therefore, to avoid the use of specific elastic bands and additional designs to eliminate finger lag, we chose this approach to help us to balance durability, affordability and grasping performance of the hand. Thirdly, this configuration intuitively demonstrates the importance of adaptive synergy between thumb and index finger in grasping tasks, as well as the soft synergy between different tendons.} 
	
	\textcolor{black}{It is straightforward to reconfigure the tendon scheme of the BPI SoftHand, although in practice we found the chosen scheme was effective. Alternatives include: 1) set of thumb, set of index and middle fingers, set of third and little finger; 2) set of thumb and index finger, set of middle finger, set of third and little finger; 3) set of thumb and index finger, set of the middle and third fingers, and set of the little finger.}
	
	The actuator is a 60W Maxon motor 347486, Planetary Gearhead GP, with encoder MR Type L, 1024 CPT, 3 Channels, with Line Driver. The output torque is between 0.75 and 4.5 Nm. The RS-232 port on the EPOS driver is connected to a control PC via an RS-232-USB converter.
	
	%%%%%%%%%%%%%%%%%%%%%%%%%%%%%%%%%%%%%%%%%%%%%%%%%%%%%%%%%%%%%%%%%%%%%%%%%%%%%%%%
	\section{Kinematics Simulation}
	The finger layout can make a huge difference on the grasping space of BPI SoftHand. Therefore, it is essential to test its grasping space through a kinematics simulation to modify the design, such as the position of the bearings and finger layout, prior to the considerable effort of fabricating the hand. In this section, we use software to simulate the movement of each joint and compare them with the results from test experiments of the hand built to these specifications.
	
	\begin{figure}[tb]
		\centering
		\includegraphics[scale=0.22,trim={0 0 0 0},clip]{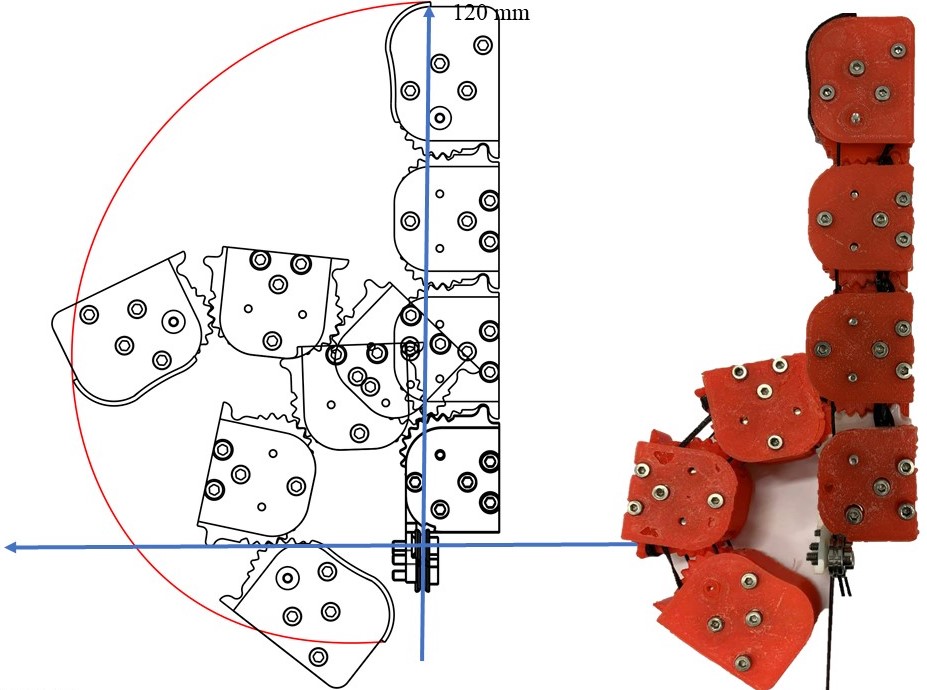}
		\caption{The left figure shows the finger kinematics simulation and the right figure shows the corresponding finger kinematics in a real experiment}
		\vspace{0em}
		\label{fig5}
		\vspace{1em}
		\centering
		\includegraphics[scale=0.25,trim={0 0 0 0},clip]{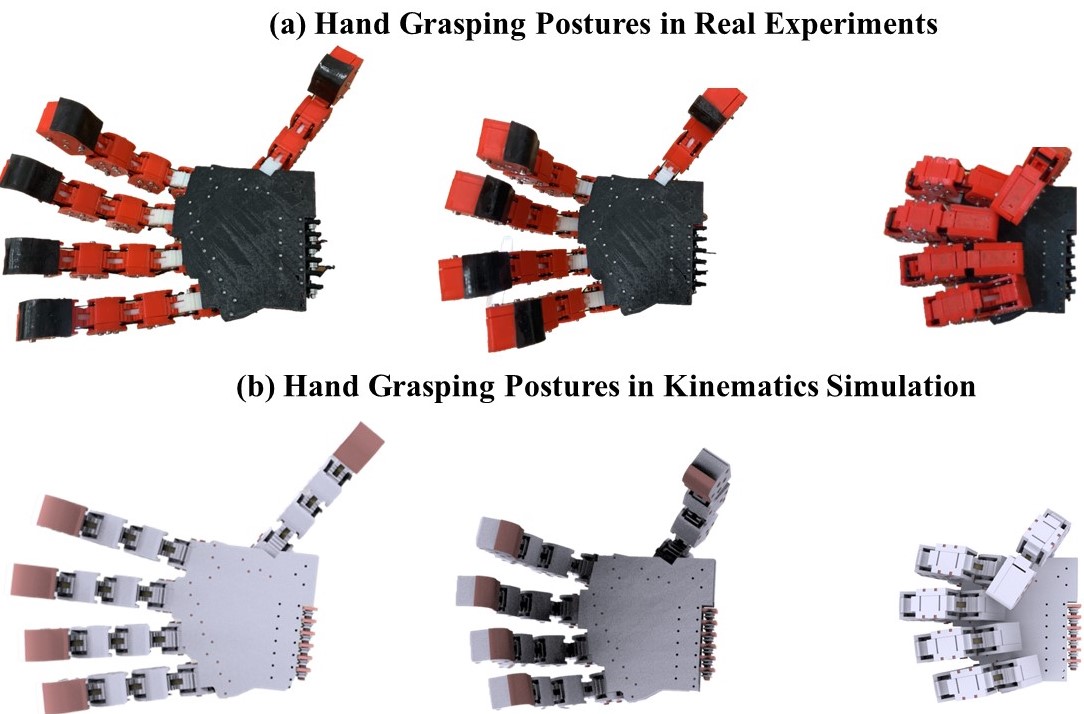}
		\caption{\textcolor{black}{(a) Hand postures in the physical experiment; (b)~hand posture in the kinematics simulation.}}
		\vspace{-1em}
		\label{fig6}
	\end{figure}
	
	Under ideal conditions, the MCP, PIP and DIP joints would rotate with the same angular speed at the same time. However, the deformation of the real tendon, and the friction between it and the U-type grooved bearings, will bring lags for the DIP and PIP joints compared with the MCP joint. Therefore, we use an empirical formula for the rotating angular relationship of these three joints, derived from the experiments with the robot fingers used here and those with the Pisa/IIT SoftHand core joint \cite{catalano2014adaptive} layout. Using $\theta_{1}, \theta_{2}, \theta_{3}$ to represent the rotation angles of the MCP, PIP and DIP joint respectively, one finds empirally that:
    $$\theta_{2}\!=\!\left (1.08+ \left | \frac{m_{1}-m_{2} }{m_{2}}\right |   \right ) \theta _{1},
    \theta _{3}\!=\!\left (\frac{\theta _{2}}{\theta _{1}}+ \left | \frac{m_{3}-m_{2} }{m_{3}}\right |   \right ) \theta _{2}$$
    
    \textcolor{black}{Then we used a Monte Carlo method to generate $\theta_{1}$ values and insert them into this empirical formula to calculate $\theta_{2}$ and $\theta_{3}$. Then, in the simulation module (Solidworks), we define the constraints between the phalanges, including gear pairs and related kinematic pairs such as rotation pairs. After entering the generated angle values, the corresponding grasping pose of the hand can then be calculated.} 
    
    \textcolor{black}{The validation involved several distinct simulations:}

	{\noindent (1) Finger kinematics simulation:} \label{sim}
	We selected representative grasping postures from the finger kinematics simulation to compare with the finger bending in a real experiment (Fig.~\ref{fig5}). The comparison figure shows that the physical rotation angles of the three joints line up well with the simulation result. This match also illustrates the accuracy of the empirical formula.
    
    \begin{figure*}[t!]
		\centering
		\begin{tabular}[b]{c}
			{\bf (a) Adaptive Synergy Experiments } \\
			\hspace{0cm}\includegraphics[width=1\textwidth,trim={0 0 0 0},clip]{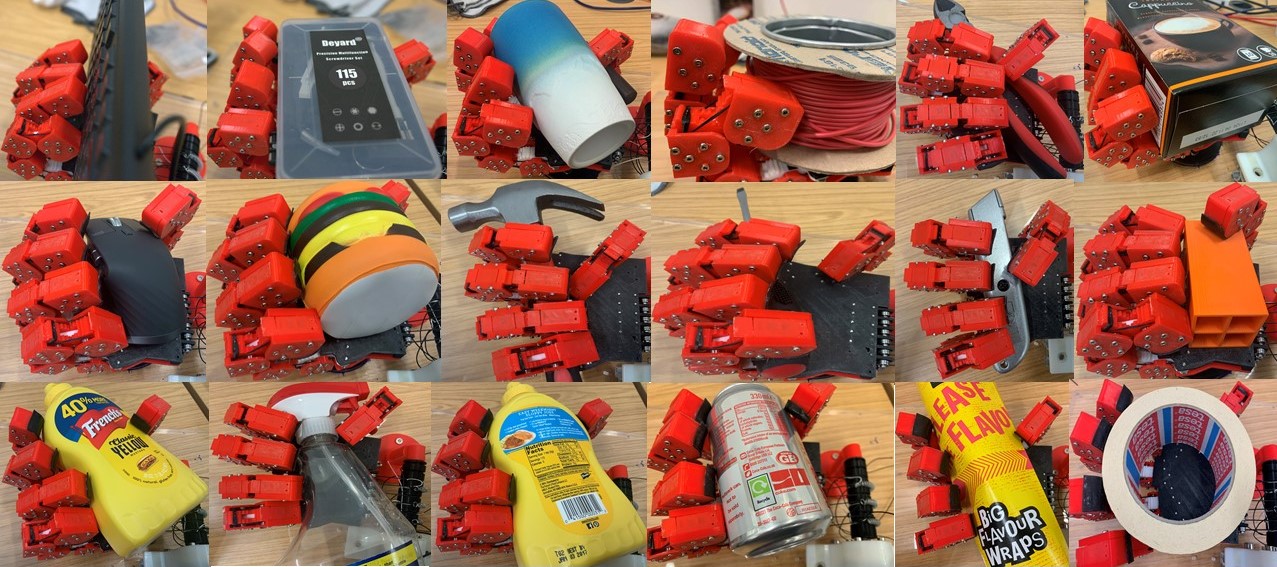} \\
		\end{tabular}
		\centering
		\begin{tabular}[b]{c}
			{\bf (b) Grasping Ability Experiments } \\
			\hspace{0cm}\includegraphics[width=1\textwidth,trim={0 0 0 0},clip]{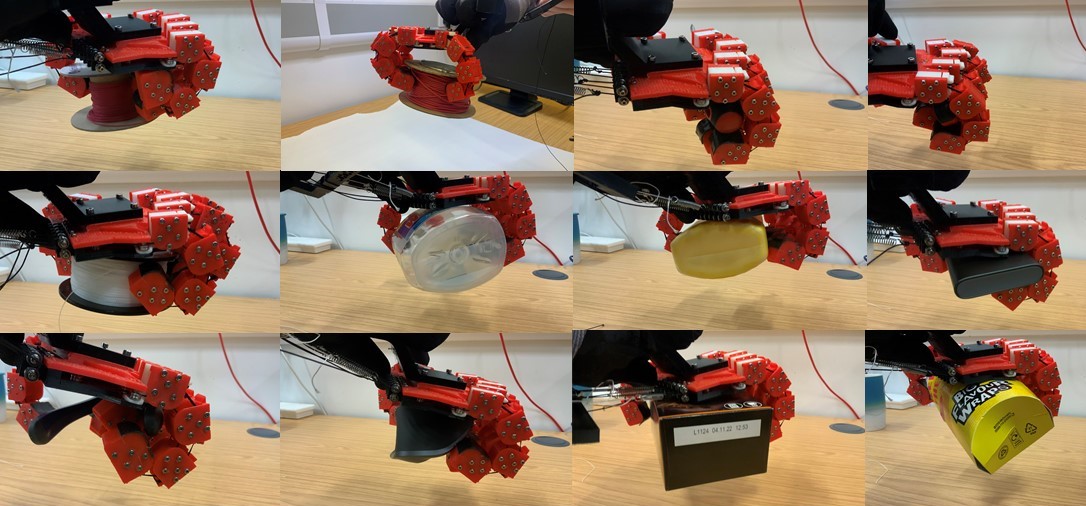} \\
		\end{tabular}
		\caption{(a) The BRL/Pisa/IIT SoftHand grasping various objects with different shapes to show its soft and adaptive synergies. (b) The BRL/Pisa/IIT SoftHand grasping various objects with different shapes to show its grasping ability. Videos of these experiments are included in the supplementary video files.}
% 		\vspace{-1em}
		\label{fig7}%
	\end{figure*}
	
    {\noindent (2) Hand-grasping kinematics simulation:}
      Again, the physical hand posture matches well with the kinematics simulation (Fig.~6), up to a small discrepancy in the rotation angles of the finger joints. Unlike the single finger test, the finger joint rotation angles are changed in the physical test due to the differing friction of the tendons and the interaction forces between the tendon and the bearings on the palm. Therefore, different tendons assign different moments to each of their corresponding joints, resulting in a slight discrepancy between the actual test and the simulation results. The simulation is sufficient, however, to tune the design of the hand to provide a suitable grasping workspace.
	 %本文实验a与Pisa/SoftHand的实验a类似。 
	%%%%%%%%%%%%%%%%%%%%%%%%%%%%%%%%%%%%%%%%%%%%%%%%%%%%%%%%%%%%%%%%%%%%%%%%%%%%%%%%	
	\section{Results} \label{Results}
	\subsection{Performance Evaluation of BRL/Pisa/IIT Hand}
	\subsubsection{Experiment A}

    This first experiment tests the synergy ability of the SoftHand in the laboratory environment. Experiment a in this paper is similar to experiment a of Pisa/SoftHand\cite{catalano2014adaptive}. The SoftHand is fixed on an acrylic plate and driven by Maxon motor. We then placed objects with various shapes such as food, kitchen cleaning items, mechanical tools and some YCB objects \cite{calli2015ycb} onto the palm of the SoftHand. By driving the Maxon motor under the speed control module of the software EPOS studio, this in turn drives the tendons to cause each joint of the robot hand to rotate, resulting on a grasp on the objects. By grasping these objects, we observe the contacts between the fingers and each object. These number-of-contacts data are counted and analyzed to verify the synergy ability of the SoftHand. In addition, a FSR (Force Sensitive Resistor) is fixed onto the hand palm, which is used to measure the holding force.

	\subsubsection{Experiment B}
    To evaluate the grasping ability of the BPI SoftHand, we perform an experiment similar to that used to evaluate Pisa/SoftHand\cite[Experiment C]{catalano2014adaptive}. The robotic hand was fixed onto an operator's arm through an auxiliary device attached to the forearm, then the robotic hand was driven manually by pulling upon a grip attached to the tendons to close the hand and thus grasp objects. The objects to be grasped are the same as those in Experiment A above, including kitchen items, common household items, tools, etc. In all cases, the operator grasped from a direction directly above the item while holding the hand perpendicular to the surface on which the object rests, to attempt to minimise the effects of human factors on grasping experiments.

	%%%%%%%%%%%%%%%%%%%%%%%%%%%%%%%%%%%%%%%%%%%%%%%%%%%%%%%%%%%%%%%%%%%%%%%%%%%%%%%%	
	\subsection{Experimental Results}
    
    \begin{figure}[t]
		\centering
		\includegraphics[scale=0.55,trim={0 0 0 0},clip]{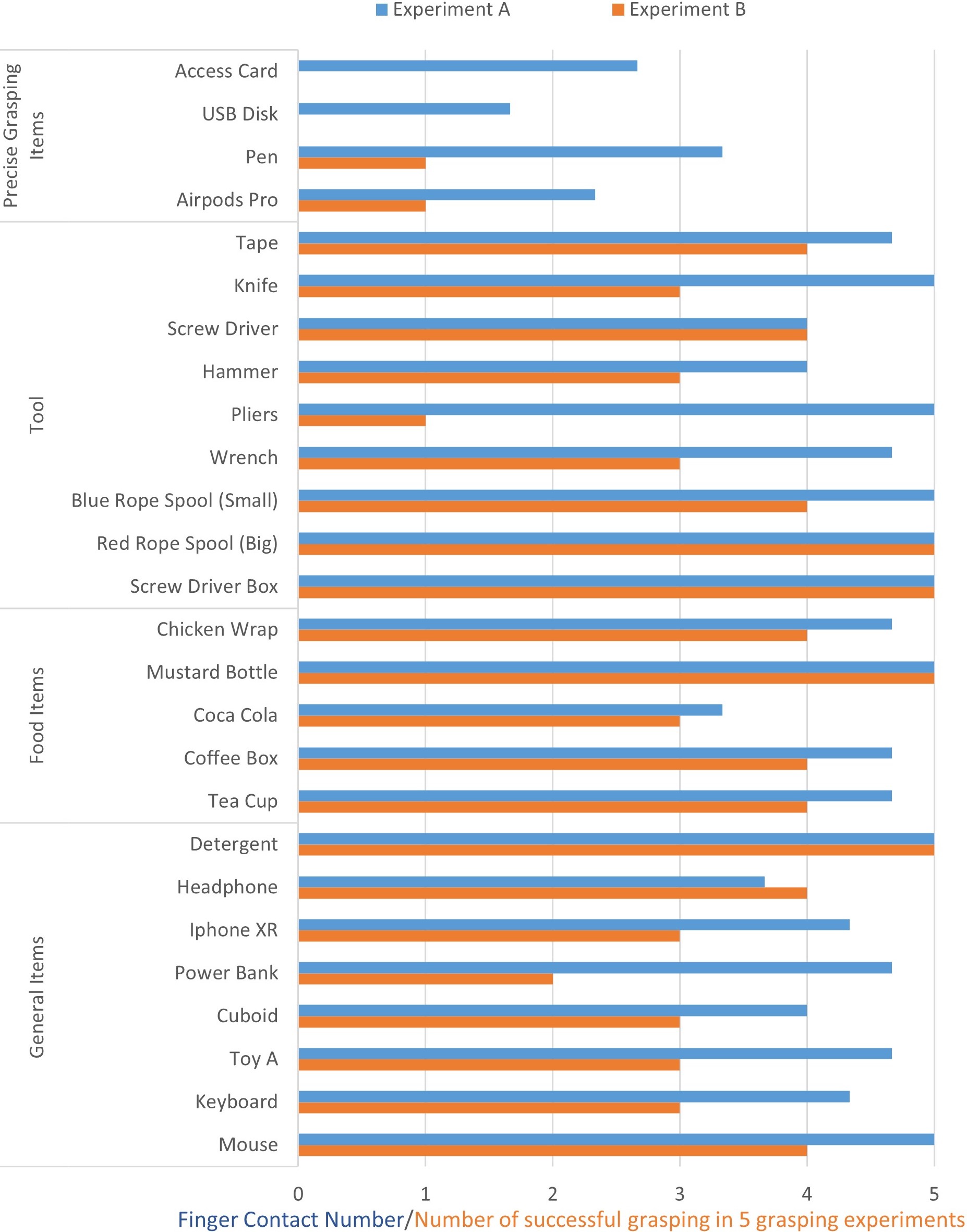}
		\caption{Statistics of Experiment A and Experiment b. The blue bar represents the number of finger contact during experiment A; The orange bar represents the number of successful grasping from 5 trials.}
		\vspace{-1em}
		\label{fig9}
	\end{figure}
	
	\begin{figure}[t]
		\centering
		\includegraphics[scale=0.6,trim={0 0 0 0},clip]{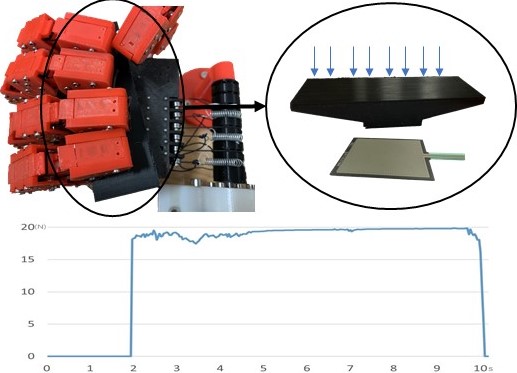}
		\caption{\textcolor{black}{The Maximum Holding Force of BPI SoftHand showing the FSR that was fixed on the palm of the hand.}}
		\vspace{-1em}
		\label{fig10}
	\end{figure}
	
	\subsubsection{Experiment A}
    The BPI SoftHand appears to exhibit good synergy in grasping many objects with various shapes, as can be seen in several images of enclosed hand around different objects (Fig.~\ref{fig7}a). For example, in grasping the spool with red-rope (3rd from right, at top), when the index finger and the middle finger have touched the surface of the object, the fingertip of the little finger can still be bent until it touches the red rope. Pictures of SoftHand gripping objects such as tape, mouse, YCB kitchen supplies also show its adaptive capabilities: in all cases the hand conforms to the various distinct shapes of these complex-shaped objects. Like the Pisa/IIT SoftHand\cite{catalano2014adaptive}, the BPI SoftHand has a good ability to adapt to object shape, and performs better in grasping objects with complex geometric shapes.
	
	For a basic yet indicative measure of grasp quality, we quantify the average number of finger contacts of the SoftHand over repeated grasps of the same object (Fig.~\ref{fig9}, blue bars; 5 grasps per object) using visual observation of the grasps. Except for some small items (e.g. airpods) and thin items (e.g. access cards), most items such as the tools and food items can be reached with 5 or 4 fingers. 
	
	In an additional experiment, the maximum holding force of the SoftHand is measured upon holding a trapezoidal object. \textcolor{black}{This experiment used a custom trapezoidal object with an embedded FSR 406 force sensor with 4.4cm$\times$3.8cm sensing area.} The finger exerts pressure on the top of trapezoidal object, which is sensed by the sensor on the bottom.  We obtained a relatively constant value of is 19.8N (see Fig.~\ref{fig10}). \textcolor{black}{We also measured the maximum press force of a single finger by placing the FSR 406 sensor on the palm underneath the position of the little (pinky) fingertip under full hand closure. The peak value 5.5N is indicative of the force from a single finger when the motor is driving the entire hand.}
	\subsubsection{Experiment B}
	The robotic hand successfully grasped most of the medium and large objects such as large spool, small spool, screw driver, chicken wrap, logitech mouse, coffee box and detergent (images shown in Fig.~\ref{fig7}b). The overall success rate is around 80\%. The relevant experimental statistics are shown in Fig.~\ref{fig9} for the various items, with the orange column represents the success rate of the SoftHand grasping objects from 5 trials. 
	
	The poorest performance of the hand was on smaller objects, such as access card and usb stick, which were not successfully grasped. Other anthropomorphic hands, included the Pisa/IIT SoftHand, also have difficult grasping such objects, and modifications (e.g. a fingernail) may be necessary for this performance. In the experiments of grasping thin objects (e.g. pliers, wrenches, iphones, keyboards), there are successful grasping cases but the average success rate is lower ($<$40\%). We attribute this to restricting the starting posture of the robotic hand grasping according to the experiment design in an attempt to standardize human factors. Another contributing reason is that the original pose of the object also affects the grasping performance, which could be improved by letting the human user `knock' the object before attempting to grasp and lift.

	%%%%%%%%%%%%%%%%%%%%%%%%%%%%%%%%%%%%%%%%%%%%%%%%%%%%%%%%%%%%%%%%%%%%%%%%%%%%%%%%	
	\section{Discussion}
    In this study, we presented a novel, low-cost, 3D-printed, underactuated and tendon-driven SoftHand. The mechanical design of its key components, such as the phalanges, can be applied to many tendon-driven underactuated robotic hands. The BRL/Pisa/IIT SoftHand is based on the transmission principles of soft and adaptive synergies. Although our SoftHand was printed with a low-precision home-use 3D printer, its synergy and gripping ability were found to be satisfactory, as verified by experiments to probe how the hand contacts the object and its performance on lifting the object.
    
    Although the BRL/Pisa/IIT SoftHand performs well overall in the experiments, in some cases such as grasping small and thin objects, the performance could be improved. Grasping small and thin objects has always been a difficult issue for robotic hands\cite{catalano2014adaptive}\cite{wang2019eagle}\cite{gao2021anthropomorphic}. In addition, the lack of feedback on the SoftHand also affected its gripping ability. At present, the motor cannot exert a different grip on different objects without having some form of sensory feedback. Fortunately, the design of the hand facilitates a high degree of user customizability, and the 3D-printed phalanges can be easily modified to explore alternative designs that may be better able to handle small objects. Another direction to take the research is to integrate tactile feedback within the hand design, such as the TacTip that has been integrated within several robot hands including the Pisa/IIT SoftHand \cite{9499032}.

    \textcolor{black}{The BRL/Pisa/IIT SoftHand will be open sourced and kept updated on GitHub (https://github.com/SoutheastWind/BPI-SoftHand), including its CAD model, the selection of components and fabrication instructions.}
    
    %{\em Acknowledgements:} We thank Andrew Stinchcombe for helping with the 3D-printing of the SoftHand.

	\bibliographystyle{IEEEtran}
	\bibliography{IEEEabrv,Refs}

\begin{thebibliography}{10}
\providecommand{\url}[1]{#1}
\csname url@samestyle\endcsname
\providecommand{\newblock}{\relax}
\providecommand{\bibinfo}[2]{#2}
\providecommand{\BIBentrySTDinterwordspacing}{\spaceskip=0pt\relax}
\providecommand{\BIBentryALTinterwordstretchfactor}{4}
\providecommand{\BIBentryALTinterwordspacing}{\spaceskip=\fontdimen2\font plus
\BIBentryALTinterwordstretchfactor\fontdimen3\font minus
  \fontdimen4\font\relax}
\providecommand{\BIBforeignlanguage}[2]{{%
\expandafter\ifx\csname l@#1\endcsname\relax
\typeout{** WARNING: IEEEtran.bst: No hyphenation pattern has been}%
\typeout{** loaded for the language `#1'. Using the pattern for}%
\typeout{** the default language instead.}%
\else
\language=\csname l@#1\endcsname
\fi
#2}}
\providecommand{\BIBdecl}{\relax}
\BIBdecl

\bibitem{bicchi2011modelling}
A.~Bicchi, M.~Gabiccini, and M.~Santello, ``Modelling natural and artificial
  hands with synergies,'' \emph{Philosophical Transactions of the Royal Society
  B: Biological Sciences}, vol. 366, no. 1581, pp. 3153--3161, 2011.

\bibitem{catalano2014adaptive}
M.~G. Catalano, G.~Grioli, E.~Farnioli, A.~Serio, C.~Piazza, and A.~Bicchi,
  ``Adaptive synergies for the design and control of the pisa/iit softhand,''
  \emph{The International Journal of Robotics Research}, vol.~33, no.~5, pp.
  768--782, 2014.

\bibitem{piazza2019century}
C.~Piazza, G.~Grioli, M.~Catalano, and A.~Bicchi, ``A century of robotic
  hands,'' \emph{Annual Review of Control, Robotics, and Autonomous Systems},
  vol.~2, pp. 1--32, 2019.

\bibitem{della2017quest}
C.~Della~Santina, C.~Piazza, G.~M. Gasparri, M.~Bonilla, M.~G. Catalano,
  G.~Grioli, M.~Garabini, and A.~Bicchi, ``The quest for natural machine
  motion: An open platform to fast-prototyping articulated soft robots,''
  \emph{IEEE Robotics \& Automation Magazine}, vol.~24, no.~1, pp. 48--56,
  2017.

\bibitem{9499032}
N.~F. Lepora, ``Soft biomimetic optical tactile sensing with the tactip: A
  review,'' \emph{IEEE Sensors Journal}, vol.~21, no.~19, pp. 21\,131--21\,143,
  2021.

\bibitem{james2021tactile}
J.~W. James, A.~Church, L.~Cramphorn, and N.~F. Lepora, ``Tactile model o:
  Fabrication and testing of a 3d-printed, three-fingered tactile robot hand,''
  \emph{Soft Robotics}, vol.~8, no.~5, pp. 594--610, 2021.

\bibitem{lepora2021towards}
N.~F. Lepora, C.~Ford, A.~Stinchcombe, A.~Brown, J.~Lloyd, M.~G. Catalano,
  M.~Bianchi, and B.~Ward-Cherrier, ``Towards integrated tactile sensorimotor
  control in anthropomorphic soft robotic hands,'' in \emph{2021 IEEE
  International Conference on Robotics and Automation (ICRA)}.\hskip 1em plus
  0.5em minus 0.4em\relax IEEE, 2021, pp. 1622--1628.

\bibitem{kappassov2015tactile}
Z.~Kappassov, J.-A. Corrales, and V.~Perdereau, ``Tactile sensing in dexterous
  robot hands,'' \emph{Robotics and Autonomous Systems}, vol.~74, pp. 195--220,
  2015.

\bibitem{butterfass2001dlr}
J.~Butterfa{\ss}, M.~Grebenstein, H.~Liu, and G.~Hirzinger, ``Dlr-hand ii: Next
  generation of a dextrous robot hand,'' in \emph{Proceedings 2001 ICRA. IEEE
  International Conference on Robotics and Automation (Cat. No. 01CH37164)},
  vol.~1.\hskip 1em plus 0.5em minus 0.4em\relax IEEE, 2001, pp. 109--114.

\bibitem{fan2018research}
S.~Fan, H.~Gu, Y.~Zhang, M.~Jin, and H.~Liu, ``Research on adaptive grasping
  with object pose uncertainty by multi-fingered robot hand,''
  \emph{International Journal of Advanced Robotic Systems}, vol.~15, no.~2, p.
  1729881418766783, 2018.

\bibitem{alspach2018design}
A.~Alspach, J.~Kim, and K.~Yamane, ``Design and fabrication of a soft robotic
  hand and arm system,'' in \emph{2018 IEEE International Conference on Soft
  Robotics (RoboSoft)}.\hskip 1em plus 0.5em minus 0.4em\relax IEEE, 2018, pp.
  369--375.

\bibitem{martin2014design}
J.~Martin and M.~Grossard, ``Design of a fully modular and backdrivable
  dexterous hand,'' \emph{The International Journal of Robotics Research},
  vol.~33, no.~5, pp. 783--798, 2014.

\bibitem{palli2014dexmart}
G.~Palli, C.~Melchiorri, G.~Vassura, U.~Scarcia, L.~Moriello, G.~Berselli,
  A.~Cavallo, G.~De~Maria, C.~Natale, S.~Pirozzi \emph{et~al.}, ``The dexmart
  hand: Mechatronic design and experimental evaluation of synergy-based control
  for human-like grasping,'' \emph{The International Journal of Robotics
  Research}, vol.~33, no.~5, pp. 799--824, 2014.

\bibitem{jiang2014modular}
L.~Jiang, B.~Zeng, S.~Fan, K.~Sun, T.~Zhang, and H.~Liu, ``A modular
  multisensory prosthetic hand,'' in \emph{2014 IEEE International Conference
  on Information and Automation (ICIA)}.\hskip 1em plus 0.5em minus 0.4em\relax
  IEEE, 2014, pp. 648--653.

\bibitem{xiong2016design}
C.-H. Xiong, W.-R. Chen, B.-Y. Sun, M.-J. Liu, S.-G. Yue, and W.-B. Chen,
  ``Design and implementation of an anthropomorphic hand for replicating human
  grasping functions,'' \emph{IEEE Transactions on Robotics}, vol.~32, no.~3,
  pp. 652--671, 2016.

\bibitem{liow2019olympic}
L.~Liow, A.~B. Clark, and N.~Rojas, ``Olympic: A modular, tendon-driven
  prosthetic hand with novel finger and wrist coupling mechanisms,'' \emph{IEEE
  Robotics and Automation Letters}, vol.~5, no.~2, pp. 299--306, 2019.

\bibitem{mizushima2018multi}
K.~Mizushima, T.~Oku, Y.~Suzuki, T.~Tsuji, and T.~Watanabe, ``Multi-fingered
  robotic hand based on hybrid mechanism of tendon-driven and jamming
  transition,'' in \emph{2018 IEEE International Conference on Soft Robotics
  (RoboSoft)}.\hskip 1em plus 0.5em minus 0.4em\relax IEEE, 2018, pp. 376--381.

\bibitem{chen2014mechanical}
W.~Chen, C.~Xiong, and S.~Yue, ``Mechanical implementation of kinematic synergy
  for continual grasping generation of anthropomorphic hand,'' \emph{IEEE/ASME
  Transactions on mechatronics}, vol.~20, no.~3, pp. 1249--1263, 2014.

\bibitem{sun2021design}
B.-Y. Sun, X.~Gong, J.~Liang, W.-B. Chen, Z.-L. Xie, C.~Liu, and C.-H. Xiong,
  ``Design principle of a dual-actuated robotic hand with anthropomorphic
  self-adaptive grasping and dexterous manipulation abilities,'' \emph{IEEE
  Transactions on Robotics}, 2021.

\bibitem{calli2015ycb}
B.~Calli, A.~Singh, A.~Walsman, S.~Srinivasa, P.~Abbeel, and A.~M. Dollar,
  ``The ycb object and model set: Towards common benchmarks for manipulation
  research,'' in \emph{2015 international conference on advanced robotics
  (ICAR)}.\hskip 1em plus 0.5em minus 0.4em\relax IEEE, 2015, pp. 510--517.

\bibitem{wang2019eagle}
T.~Wang, Z.~Geng, B.~Kang, and X.~Luo, ``Eagle shoal: A new designed modular
  tactile sensing dexterous hand for domestic service robots,'' in \emph{2019
  International Conference on Robotics and Automation (ICRA)}.\hskip 1em plus
  0.5em minus 0.4em\relax IEEE, 2019, pp. 9087--9093.

\bibitem{gao2021anthropomorphic}
G.~Gao, A.~Dwivedi, and M.~Liarokapis, ``An anthropomorphic prosthetic hand
  with an active, selectively lockable differential mechanism: Towards
  affordable dexterity,'' in \emph{2021 IEEE/RSJ International Conference on
  Intelligent Robots and Systems (IROS)}.\hskip 1em plus 0.5em minus
  0.4em\relax IEEE, 2021, pp. 6147--6152.

\end{thebibliography}
	
\end{document}